\documentclass{article}

\usepackage{arxiv}

\usepackage[utf8]{inputenc} 
\usepackage[T1]{fontenc}    
\usepackage{hyperref}       
\usepackage{url}            
\usepackage{booktabs}       
\usepackage{amsfonts}       
\usepackage{nicefrac}       
\usepackage{microtype}      
\usepackage{lipsum}		
\usepackage{graphicx}
\usepackage{subcaption}

\usepackage{doi}

\title{Optimal latent space forecasting for large collections of short time series using Temporal Matrix Factorization}

\author{{\hspace{1mm}Himanshi Charotia}\\
	Mastercard AI Garage\\
	Gurgaon, India\\
	\And
	{\hspace{1mm}Abhishek Garg} \\
	Mastercard AI Garage\\
	Gurgaon, India\\
	\And
	{\hspace{1mm}Gaurav Dhama} \\
	Mastercard AI Garage\\
	Gurgaon, India\\
	\And
	{\hspace{1mm}Naman Maheshwari} \\
	Mastercard AI Garage\\
	Gurgaon, India\\
}

\renewcommand{\shorttitle}{Optimal Latent Space Forecasting}

\hypersetup{
pdftitle={A template for the arxiv style},
pdfsubject={q-bio.NC, q-bio.QM},
pdfauthor={David S.~Hippocampus, Elias D.~Striatum},
pdfkeywords={First keyword, Second keyword, More},
}

\begin{document}
\maketitle

\begin{abstract}
In the context of time series forecasting, it is a common practice to evaluate multiple methods and choose one of these methods or an ensemble for producing the best forecasts. However, choosing among different ensembles over multiple methods remains a challenging task that undergoes a combinatorial explosion as the number of methods increases. In the context of demand forecasting or revenue forecasting, this challenge is further exacerbated by a large number of time series as well as limited historical data points available due to changing business context. Although deep learning forecasting methods aim to simultaneously forecast large collections of time series, they become challenging to apply in such scenarios due to the limited history available and might not yield desirable results. We propose a framework for forecasting short high-dimensional time series data by combining low-rank temporal matrix factorization and optimal model selection on latent time series using cross-validation. We demonstrate that forecasting the latent factors leads to significant performance gains as compared to directly applying different uni-variate models on time series. Performance has been validated on a truncated version of the M4 monthly dataset which contains time series data from multiple domains showing the general applicability of the method. Moreover, it is amenable to incorporating the analyst view of the future owing to the low number of latent factors which is usually impractical when applying forecasting methods directly to high dimensional datasets.
\end{abstract}


\section{Introduction}
With the easy availability of high-end computing, modern business forecasting applications typically forecast millions of time series depending on the business requirements. These forecasts might be required for different time horizons (monthly, quarterly or yearly) depending on the particular business context and the accuracy of certain time series predictions might matter more than the others (e.g. company revenue predictions for external investor relations vs product level revenues for internal financial planning). One of the most crucial decisions that determine the successful prediction of these series is selecting the appropriate forecasting method which is best suited for each series depending upon its properties (trend, seasonality, etc.). For high-dimensional time series, the analyst typically needs to resort to multivariate methods or design a system that can automatically select the most appropriate uni-variate method. The complexity further adds up when the time series is short in length since historical data beyond a certain point might not be relevant because of constantly changing business scenarios.

As observed in many real-life scenarios, there is often a need to forecast demand for products/customers for which no/less history is present. This can arise due to lack of good data collection method (rainfall data may not be present due to infrastructure limitation~\cite{rainfall}), a new product launch (every season the substantial amount of new products are launched in fashion industry~\cite{fashion_attention_paper}) or due changing business scenarios(computing the capital gain and profit for risk management \cite{Indonesia_stock_exchange}). And short term time series come with their own challenges: lack of history for the model to estimate parameters from and amount of randomness in the data.

It is very common for extremely simple forecasting methods like "forecast the historical average" to outperform more complex methods in the case of short term time series. Using least squares estimation or some other non-regularized estimation method, it is possible to estimate a model only if you have more observations than parameters \cite{short_series_rob_hyndman}. Complex methods like Deep Learning are naturally omitted due to a large number of parameters to be optimized in them.  

The only reasonable approach is to first check that there are enough observations to estimate the model, and then to test if the model performs well out-of-sample. With most business scenarios, sufficient data points may not be present to conclude. AIC can not be used in this case as it suggests very simple models because any model with more than one or two parameters will produce poor forecasts due to the estimation error. Further, \cite{3b1355aedd1041f1853e609a410576f3} suggested the use of time series cross-validation in which models from many different classes may be applied, and the model with the lowest cross-validated MSE  selected. However, this method is not efficient as the computation time involved increases quadratically with the number of series to be forecasted. So for the datasets containing short-length high dimensional time series two dominant problems needs to be solved: 1) optimal model selection with a short history and 2) scaling of the solution to a large collection of time series.  

For forecasting a large collection of time series, several notable approaches have been proposed in the literature based on matrix/tensor factorization\cite{xiong2010temporal,NMF}. Matrix factorization decomposes the time series into a small set of latent factors based on the assumption that series in a dataset exhibit high correlation and shared latent properties. As Matrix/tensor factorization scales linearly with \textit{n}, it presents a natural solution to address the scalability and efficiency issue but there is no factor to capture temporal dependency in them. This leads to latent factors being highly erratic and with a low signal-to-noise ratio.
To counter this, \cite{NIPS} proposed the Temporal Regularized Matrix Factorization (TRMF) scheme to model multivariate time series by introducing a novel AR regularization scheme on the temporal factor matrix. This work was further extended by \cite{BTMF} to provide probabilistic forecasts and incorporated a vector autoregressive(VAR) to process the temporal factor matrix. Authors in \cite{BTMF} argued that although TRMF proposes a superior factorization scheme it fails to take into account any dependencies between the latent factors.

Overall, these factorization approaches have shown superior performance in modeling real-world large-scale time series data; however, these methods limit the latent factor forecasts to simple models such as simple AR or VAR processes limiting their application. These models in general require careful tuning of the regularization parameters to ensure model accuracy and BTMF requires sufficient history to model VAR process thus proving inefficient for modeling short time series. 

We seek to address this problem by proposing a new framework for modeling short length large-scale time series by first decomposing series into latent vectors using the TRMF scheme and then selecting the most optimal method from a wide variety of univariate methods using cross-validation. As each of the individual time series is a linear combination of these latent factors, the decomposition allows for an optimal combination of different forecasting methods depending on the weight of these latent factors in each time series.

The rest of this paper is organized as follows. We review the existing approaches for analyzing large-scale time series and optimal model selection in Section 2. In Section 3, we explain the detailed components and procedures of our proposed framework for forecast-model selection. Section 4 describes the data and methods used for the experimentation. We demonstrate the superiority of the proposed approach via extensive experimental results in Section 5 and conclude the paper in Section 6.


\section{Related Work} 

\subsection{Factorizing Time series}
In recent literature, a lot of work has been done by applying matrix factorization (collaborative filtering) to analyze large-scale time series. The central challenge that this approach faces is to incorporate temporal dynamics in the learned embeddings by designing appropriate regularization terms, achieving high accuracy and overfitting. In standard matrix factorization, the Frobenius norm is used as a regularizer, which can not be used in data with temporal dependencies as it is invariant to column permutation. 
Some of the existing temporal MF approaches turn to the framework of graph-based regularization \cite{smola2003kernels}, with a graph encoding the temporal dependencies. Authors in \cite{sun2014collaborative} presented a dynamic matrix factorization model using collaborative Kalman filtering to address the scalability issue in linear dynamical systems for large-scale time series.
Authors in \cite{NIPS} used a novel AR regularization scheme on the temporal factor matrix. \cite{takeuchi2017autoregressive} extended \cite{NIPS} and introduced graph Laplacian regularizer to model spatial correlations in tensor data along with temporal dynamics. Authors in \cite{BTMF} integrated VAR and matrix/tensor factorization into a single probabilistic framework to model large-scale time series. These matrix/tensor factorization-based algorithms seem like a natural solution to model large-scale time series data. It helps in uncovering latent temporal patterns in data and is scalable.

\subsection{Optimal Model selection}
One of the challenging questions in time series forecasting is how to find the best algorithm. The two most basic approaches for selecting the forecasting method are aggregate selection and individual selection \cite{article}. For individual selection, two major approaches used in literature are information criteria (AIC) and empirical accuracy. For evaluating empirical accuracy cross-validation and error measures are used\cite{HYNDMAN2006679}. All these approaches lack in terms of computational accuracy in case of the short length of series and are not scalable. Another type of approach which is deployed is the use of a rule based system which suggests a forecasting method for a particular time series based on its statistical characteristics \cite{ARINZE1994647,https://doi.org/10.1002/1099-131X(200011)19:6<515::AID-FOR754>3.0.CO;2-7,WANG20092581}. In \cite{widodo2013model}  authors first applied principal component analysis to reduce the dimension of their data and then proposed a model to select the proper forecasting method. Authors in \cite{kuck2016meta}, studied how the different feature sets including past forecast error-based features and statistical tests impact the performance of neural network as a meta-learner to select among different forecasting  models. Authors in \cite{talagala2018meta}, proposed a general framework in which they used time series features as meta-features and selected the best forecasting algorithm for time series using the random forest classifier. Regardless of many studies performed in this field, not much emphasis has been given to short length time series.

\section{Proposed Method }
The main goal is to provide automated forecasts for short length large-scale time series by a framework that selects the best method for each time series without going into the tedious task of handpicking the forecasting method. To address this problem, Section 3.1 and 3.2 discuss the proposed framework in detail which would automatically select the best forecast for individual series using matrix factorization and cross-validation. High-dimensional time series data can be represented in the form of a matrix. Consider a High-dimensional time series denoted by matrix \( Y \in \mathbb{R}^{N\times T} \) collected for \textit{N} sources on \textit{T} time stamps, with each row \( y_i = (y_{i,1}, y_{i,2},...., y_{i,t}, y_{i,t+1},...., y_{i,T} )\) corresponding to the time series collected at source \textit{i} for \textit{T} time steps. Each \(y_{i,T} \) can be produced by taking inner product of \(\textbf{f}_i^T\textbf{x}_t \) by following the general matrix factorization model where \(\textbf{f}_i \in \mathbb{R}^K\) is \textit{K}-dimensional latent factor matrix for \textit{i}-th time series and \(\textbf{x}_t \in \mathbb{R}^K\) is \textit{K}-dimensional temporal factor matrix for \textit{t}-th time point. We can stack \( \textbf{x}_{t}\) to form column vector \( X \in \mathbb{R}^{K\times T}  \)
and \( \textbf{f}_{i}^T\) to form row vector \( F \in \mathbb{R}^{K\times N}  \)
and
\begin{equation}
 Y \approx F^T X 
\end{equation}
where \textit{X} is temporal embedding matrix for each time step and each time series \(y_i \in Y \)  can be reconstructed from this base \textit{X} matrix using coefficients given by \textit{F}. One can predict \( \textbf{x}_{t+1} \) on the latent temporal factor matrix \textit{X} by training individual models for \textit{K} time series as \(\textit{K}\ll\textit{N}\) and then can get estimated time series data at \textit{t}+1 with \( y_{i,t+1} \approx \textbf{f}_i^T \textbf{x}_{t+1} \)

Selecting optimal methods for \textit{K} time series is done by first training different families of forecasting methods given in Table \ref{table : model_parameters} and by ranking these methods based on evaluation metrics (sMAPE) by observing performance on fitted value using nested cross-validation. Then the final forecast is provided as the median of the top three ranked methods.

\subsection{Latent Representation}
The first part of the framework is to represent high-dimensional time series as a small set of latent factors. For decomposing the time series into its latent representation, we use the decomposition performed by TRMF \cite{NIPS} which not only generates a latent representation of time series but also encourages temporal structure among its generated embedding. TRMF tries to enforce temporal structure into its embedding by use of regularizers for both \textit{F} and \textit{X}. Matrix factorization in TRMF can be represented by the equation below \cite{NIPS}:
\begin{equation}
\min_{F,X}\sum_{(i,t)\in \Omega} (Y_{it}-f_i^T x_t)^2+\lambda_f\mathcal{R}_f(F) +\lambda_x\mathcal{R}_x(X)
\end{equation}
where \(\Omega\) set of all \textit{i} and \textit{t}. \(\mathcal{R}_f(F) ,\mathcal{R}_x(X) \) are autoregressive regularizers for \textit{F} and \textit{X} respectively. TRMF takes as hyper parameter a couple of terms and \textit{length of latent embedding dimension} is one of the major one.
For selecting the optimal value of \textit{length of latent embedding dimension}, we have used elbow method \cite{elbow} as a heuristic.  The Elbow method is extensively used for selecting the optimal number of clusters in k-means. In this method, the distortion score is computed by varying the value of the number of clusters and is plotted against the number of clusters. The point of inflection(elbow) on the curve is the best value of \textit{k} (number of clusters). Similarly, we analyzed the reconstruction error of forming back \textit{Y} from \(F^TX\) as a function of the \textit{length of latent embedding dimension} and picked the elbow of the curve as the \textit{length of latent embedding dimension}. As clearly given in Figure 1, we have chosen 18 as \textit{length of latent embedding dimension} as the point of inflection is between 12-20.
\begin{figure}
\centering
\includegraphics[scale=0.6]{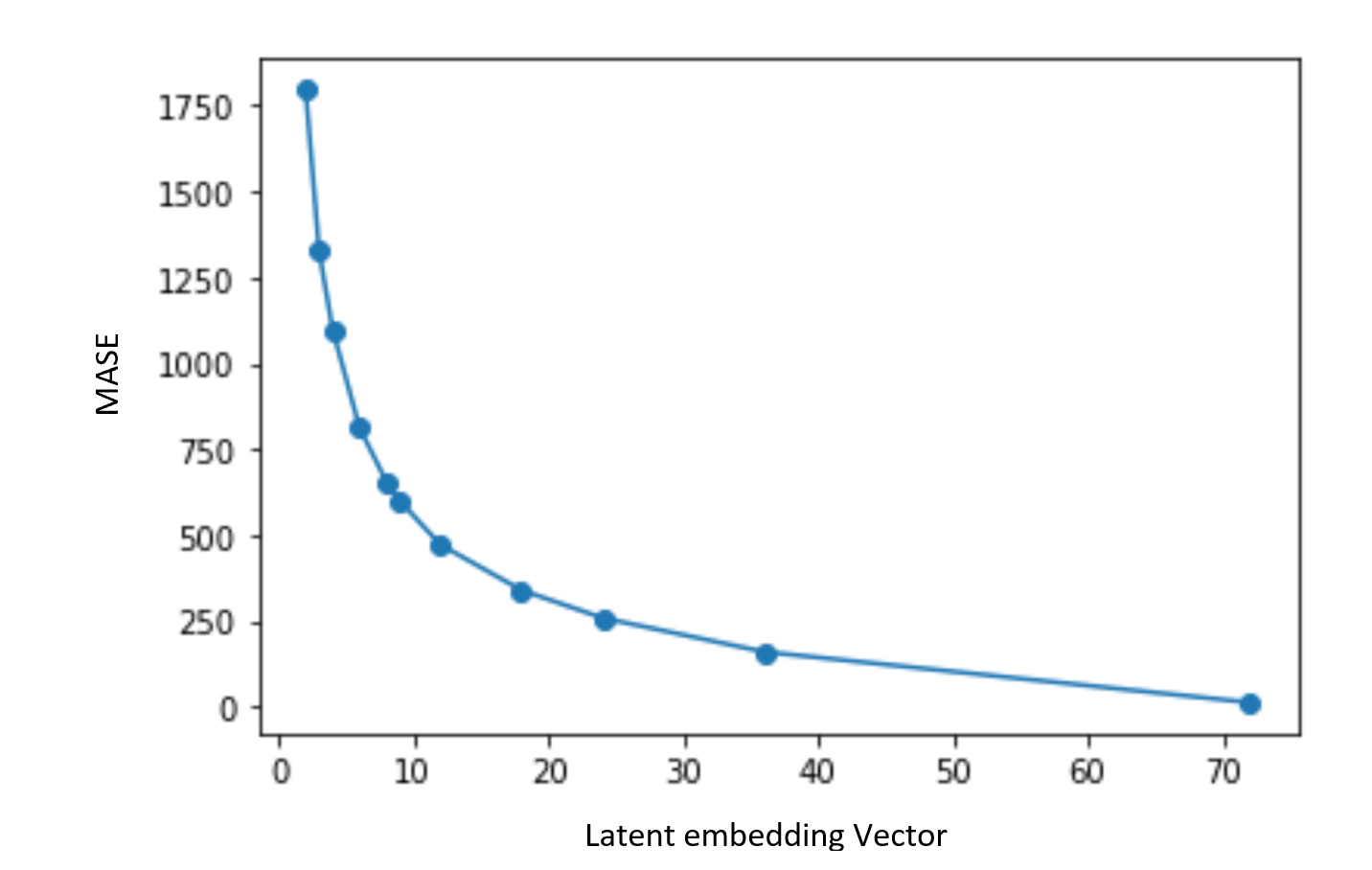}
\caption{Plot of optimal latent embedding dimension length selection with respect to reconstruction error (MASE) using elbow method.}
\end{figure}
\subsection{Optimal Method Selection}
For selecting the best method for each series in the temporal embedding matrix, we use the Forward-Chaining type of cross-validation \cite{cv} in which we divided the entire training set(60 data points) into different sets of 6 months each. We successively consider every 6 months as the test set and assign all previous data into the training set. This method produces many different train/validation splits. For each training period, we run the candidate models mentioned in Table \ref{table : model_parameters} to generate forecasts. The error on each split is averaged to compute a robust estimate of the model error. Average error (sMAPE) by each method on different sets of fitted values is used to rank these methods. The final forecast is the median of the top three ranked methods.

\section{Experimentation}
\subsection{Dataset}
The M4-Competition\cite{m4} was organized by Spyros Makridakis with the purpose to identify the most accurate forecasting method(s) for different types of time series. The M4 dataset contains time series, built from multiple, diverse, and publicly accessible sources. These series are then scaled and anonymized. The original M4 monthly dataset contained 72 points(6 years) and the next 18 months were to be predicted as part of the competition. For this paper, we have trimmed this dataset down to a maximum of 60 latest data points for training and used the later 12 points as a test period.
This step was done to make the dataset resemble a real business forecasting problem as in most business scenarios only 4-5 years of data or even less about a product is present/relevant. Table~\ref{label_category} contains the category-wise details of the monthly series of the M4 dataset. M4 monthly dataset contains most of the series coming from the Finance and Micro industry making it well suited for the proposed problem. It also contains time series from other domains such as Demographic, Macro, etc which supports generalization of results on other domains as well.

\subsection{Methods}
\begin{figure*}
\centering
\includegraphics[width=16cm,height=11cm]{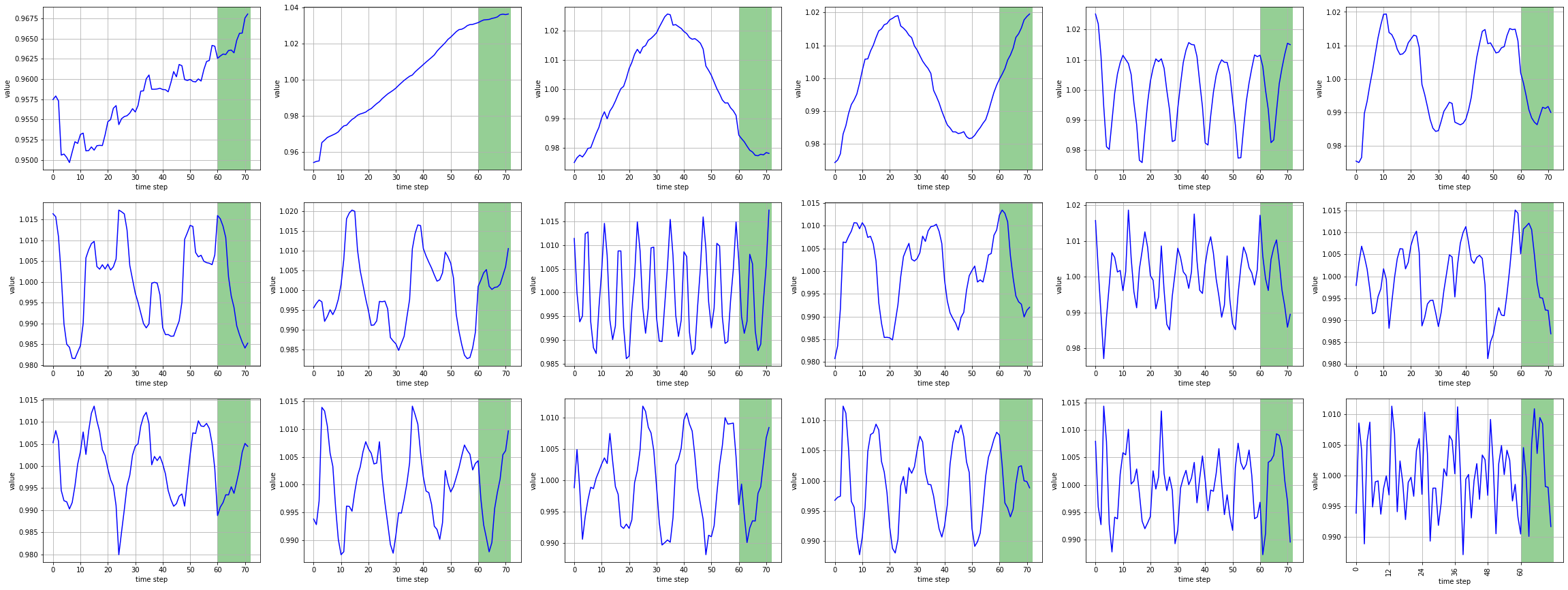}
\caption{ Plot of latent embedding of time series present in trimmed version of M4 monthly dataset generated by TRMF. Shaded region denotes forecasting period.}\label{table : model_parameters}
\end{figure*}

After the dataset has been decomposed into latent series, we compute the optimal method for each series using the methods given in Table~\ref{table : model_parameters}. We have included methods to incorporate all types of time series forecasting methods including classical time series forecasting methods (ARIMA), Hierarchical models (THIEF and MAPA), seasonal and trend decomposition models(STLM, STLM with ARIMA ), generalized models (GUM), and Deep Neural Network based models (NNETAR). The purpose of this step was to enable the framework to use forecasting method which is best suitable for individual time series instead of implementing a fixed set of method which may not be best suitable for patterns present in the series (for e.g ARIMA falls short to capture multiple seasonal patterns). We have chosen 1) TRMF \cite{NIPS} and 2) BTMF (Bayesian Temporal Factorization for
Multidimensional Time Series Prediction) \cite{BTMF} as main benchmark models. We also consider other deep neural-network based frameworks such as 3) Exponential Smoothing-Recurrent Neural Networks (ESRNN ) \cite{esrnn} which combines Exponential Smoothing model (ES) and a Recurrent Neural Network (RNN) and is the winner of the M4 competition, 4) DeepAR -Probabilistic Forecasting with Autoregressive Recurrent Networks \cite{deepar}, 5) N-BEATS \cite{nbeats} which is pure DL model for time series forecasting and 6) a simple Feed-forward neural network. These models are chosen because they have shown superior performance in tasks of optimal time series forecasting and are scalable also. All methods apart from DL frameworks and TRMF are implemented in the R language. For implementing Deep Learning models including DeepAR, N-BEATS and feedforward neural network GluonTS toolkit \cite{gluonTS} has been used. In addition, for every studied Deep learning model and TRMF, we also carried out hyper-parameter tuning wrt. n\_components, n\_order, batch size, epoch, learning rate, and context length to arrive at the right set of training parameters. In the case of ESRNN, we have used the winning model parameters in the M4 competition. 
\begin{table}
\begin{center}
\caption{Category wise distribution in M4 monthly dataset }\label{label_category}
 \begin{tabular}{|c|c|c|} 
 \hline
 Index & Type & Count \\  
 \hline\
 1 & Demographic & 5728 \\ 
 \hline
 2 & Finance & 10987 \\
 \hline
 3 & Industry & 10017  \\
 \hline
 4 & Macro & 10016  \\
 \hline
 5 & Micro & 10975\\ 
 \hline
 6 & Other & 277\\
 \hline
\end{tabular}
\end{center}
\end{table}
   
\subsection{Evaluation Metric}
To evaluate the model performance, we have used the Overall Weighted Average (OWA) proposed for the M4 competition. This metric is calculated by obtaining the average of the symmetric mean absolute percentage error (sMAPE) \cite{smape} and the mean absolute scaled error (MASE) \cite{mase} for all the time series and also calculating it for the Naive2 \cite{naive2_wheelwright1998forecasting} predictions. The Naive2 model automatically adapts to the potential seasonality of a series based on an autocorrelation test. If the series is seasonal the model composes the predictions of Naive and Seasonal Naive, else the model predicts on the simple Naive. These measurements are calculated as follows:

\begin{equation}
sMAPE = \frac{2}{h}{\sum_{t=n+1}^{n+h}\frac{|Y_t-\hat{Y_{t}}|}{|Y_t+\hat{Y_t}|} \ast 100 (\%)  },
\end{equation}

\begin{equation}
MASE = \frac{1}{h}\frac{{\sum_{t=n+1}^{n+h}{|Y_t-\hat{Y_{t}}|} }}{\frac{1}{n-m}{\sum_{t=m+1}^{n}{|Y_t-Y_{t-m}|} }},
\end{equation}
where \( Y_{t}\) is the value of the time series at point \textit{t}, \(\hat{Y_t}\) the estimated forecast, \textit{h} the forecasting horizon, \textit{n} the number of the data points available in-sample, and \textit{m} the time interval between successive observations considered by the organizers in M4 competition, i.e.,  12 for monthly frequency. Then the OWA is calculated as: 
\begin{equation}
OWA =  \frac{1}{2} {\left(\frac{sMAPE}{sMAPE_{Naive2}} + \frac{MASE}{MASE_{Naive2}} \right)},
\end{equation}

\section{Results and Analysis}
In this section, we empirically validate the framework described in Section 3. Figure 2 shows the latent embeddings collectively formed for each time series in the M4 monthly dataset. As shown in Figure 2, TRMF characterizes the temporal trend among time series very well containing varying trends (increasing and decreasing) and different types of seasonality including short-term and long-term. For comparing the proposed framework, we have chosen the most recent methods present in the literature that deals with forecasting large-scale time series namely TRMF and BTMF. Table \ref{table : label_results }  show the performance of competing models on individual series and their latent embeddings. Results by BTMF were exploding due to the short length of series as parameters in the VAR process need long history to be optimized. And therefore are not included in the final performance. Essentially, the proposed framework achieves competitive forecasting results among matrix factorization and deep learning models and suggests smaller OWA in the prediction task of short length high dimensional time series.
\begin{table}
\begin{center}
\caption{Performance when different models are applied on individual series and latent series in terms of OWA }\label{table : label_results }

\small
 \begin{tabular}{p{4cm} ||*{3}{c} ||*{3}{c}}
 \hline
\hline
\textbf \textbf{}& \multicolumn{3}{c||}{\textbf{Individual series}}
    & \multicolumn{3}{c}{\textbf{Latent series }}\\
 \hline
 \textbf{Method} & \textbf{sMAPE} & \textbf{MASE}& \textbf{OWA}& \textbf{sMAPE} & \textbf{MASE}& \textbf{OWA}\\ [1ex]
 \hline
 CV+median(Proposed method)& - &- &- &\textbf{8.22} &\textbf{ 0.49} &\textbf{0.58}   \\ 
 CV  & - &- &- &9.12& 0.56&0.65     \\ 
 \hline
  TRMF &12.52 &0.85 & 0.95  & - &- &- \\
 ESRNN &12.23 &0.93 & 0.97 & 48.72 &4.65 &4.42   \\
 DeepAR &12.97 &1.35 & 1.24  & 21.15&1.22 &  1.48\\
 N-BEATS &12.54 &0.91 & 0.97 & 18.30&1.28 &1.40 \\
  FEEDFORWARD-NN&15.32 &1.91 & 1.64  &13.70 &0.97 & 1.05\\
 \hline
 AUTO-ARIMA &13.12 &0.86& 0.97 &8.73 &0.50& 0.61 \\
 THIEF &12.47 &0.85 & 0.94 & 10.49 &0.63 & 0.75\\
 MAPA & 12.12&0.83 & 0.92  &10.20 &0.62 &0.73 \\
 NNETAR&15.49& 1.06&1.17  &10.71&0.66 &0.77  \\
  NNETAR (BOXCOX)& 15.49& 1.06 &1.17  &10.18&0.64 &0.74 \\
 GUM &13.75 &0.92& 1.03  &10.14 &0.59 &0.71\\
 STLM-AR & 16.44& 1.13& 1.25 & 13.11&0.89 &0.99 \\
 STLM (BOXCOX) & 16.44& 1.13&1.25   &11.04 & 0.64& 0.77 \\
 \hline

\end{tabular}
\end{center}


\end{table}
\subsection{Comparing model performance on individual series  }
The results in Table \ref{table : label_results } reveal that prediction via multiple models over temporal matrix factorization is more superior to predicting latent embeddings only with the AR model( in case of TRMF) or VAR model (in case of BTMF). Deep learning models like ESRNN, DeepAR, and  N-BEATS have shown state-of-the-art performance in various time series prediction tasks including M4 competition but perform worse in the case of short length time series. Essentially, the statistical models like ARIMA, THIEF, NNETAR, etc perform the worst as the temporal trend among series is ignored and a single model is rarely able to capture properties of different time series. 
\subsection{Comparing model performance on latent series }
Table \ref{table : label_results } also depicts the results achieved based on OWA when competing models are applied to latent embeddings produced by TRMF.  Practically, the Deep learning models perform the worst as they require a large number of examples for training and only 18 series were present for these models to learn. The proposed framework beats individual methods applied on latent series in terms of OWA due to varying characteristics of latent series and the inability of a single model to capture them. NNETAR being a neural network model can capture nonlinearity in the latent vectors but still lacks from a method like AUTO ARIMA as sufficient data points are required for a neural network to learn. Model selection using cross-validation does improve the overall OWA. Our results suggest that the Proposed framework inherits the advantages of matrix models (TRMF) and inherits characteristics of statistical models. It not only provides a flexible
and automatic technique for model selection but also offers superior prediction performance by integrating temporal dynamics into matrix factorization. And as the \(\textit{K}\ll\textit{N}\), the proposed framework does provide the ability to analyze each series in the latent dimension.

\section{Conclusion}
In this paper, we present a framework for optimal method selection in short length large scale time-series by incorporating Matrix Factorization and the selection of forecast method based on nested cross-validation. The integration allows us to better model the complex temporal dynamics in large-scale time series and provides a fast automated framework that is scalable also. Factorization by TRMF allows us to get latent vector embeddings per dataset which captures temporal dynamics and provides us with a dataset that is much lower in dimension and can easily be forecasted. Model selection by nested cross-validation helps to automate the model selection procedure instead of manually selecting methods for each time series. We examined the framework on the truncated version M4 monthly dataset which contains time series of varying length from different business domains and the Proposed framework has demonstrated superior performance over other baseline models.  
%
%
%
\clearpage
.\bibliographystyle{unsrt}
\bibliography{bibfile}

\clearpage
\appendix

\section{Supplement}

\begin{table}[!htbp]
\caption{Methods used for Optimal model selection. Bracket denotes the additional transformation applied on dataset. }

\begin{center}
 \begin{tabular}{|p{3cm}|p{4.5cm}|p{4.5cm}|} 
 \hline
 \textbf{Method} & \textbf{Description} & \textbf{Parameters}\\ [1ex]
 \hline
Auto Arima&Fit Best ARIMA Model To Univariate Time Series&max.p=5, max.q = 5, max.d = 2, max.P = 2, max.Q=2, max.D = 1   \\ 
 \hline
NNETAR &Feed-forward neural networks with a single hidden layer and lagged inputs for forecasting univariate time series &P=1, size=9  \\
 \hline
NNETAR(BOXCOX) &Feed-forward neural networks with a single hidden layer and lagged inputs for forecasting univariate time series &P=1, size=9, lambda="auto"  with BOXCOX transformation\\
 \hline
STLM & Applies an STL decomposition and models the seasonally adjusted data & default \\
 \hline
STLM(BOXCOX) &Applies an STL decomposition and models the seasonally adjusted data& lambda='Auto'  with BOXCOX transformation\\
 \hline
MAPA & Multiple Aggregation Prediction Algorithm& default\\
 \hline
THIEF &Temporal HIErarchical Forecasting & default\\
 \hline
GUM &Generalized Univariate Model & default\\
 \hline
 TRMF &Temporal regularized Matrix Factorization & n$\_$components = 18, n$\_$order = 6, C$\_$Z= \(5e^{1}\), C$\_$F = \(5e^{-4}\), C$\_$phi = \(1e^{-4}\), eta$\_$Z = 0.25, C$\_$B = 0, fit$\_$regression = False, fit$\_$intercept = False, nonnegative$\_$factors = False, n$\_$max$\_$iterations = 1000, tol= \(1e^{-5}\)\\
 \hline
 ESRNN & Exponential Smoothing Recurrent Neural Network- a multivariate hybrid ML(Deep Learning)-time series model &  input-size=12, output-size=12, state-hsize=50, batch-size=256, learning-rate=\(5e^{-4}\), dilations=[1,3,6,12]\\
 \hline
DeepAR &Probabilistic Forecasting with Autoregressive Recurrent Networks& context-length=24, epochs=30, learning-rate=\( 1e^{-3}\), num-batches-per-epoch = 1000\\
 \hline
N-BEATS &Neural Basis Expansion Analysis for interpretable Time Series &num-stacks=2, stack-types = ["T","S"], num-blocks = [3], widths=[256,2048], sharing = [True], expansion-coefficient-lengths = [32], loss-function = 'MASE',  epochs=30, learning$\_$rate=\(1e^{-3}\), num-batches-per-epoch = 50\\
 \hline
FeedForward-NN & Simple feedforward neural network& hidden-dimensions=[30], context-length=24, epochs=50,                      learning$\_$rate=\( 1e^{-3}\), num-batches-per-epoch = 1000\\
\hline
\end{tabular}
\end{center}

\end{table}

\end{document}